\documentclass{article}
\usepackage{spconf,amsmath,epsfig}
\usepackage{amssymb}
\let\OLDthebibliography\thebibliography
\renewcommand\thebibliography[1]{
  \OLDthebibliography{#1}
  \setlength{\parskip}{0pt}
  \setlength{\itemsep}{0pt plus 0.3ex}
}

\begin{document}\sloppy

\def\x{{\mathbf x}}
\def\L{{\cal L}}

\title{MPN: Multimodal Parallel Network for Audio-Visual Event Localization}
\name{Jiashuo Yu\textsuperscript{1}, Ying Cheng\textsuperscript{2}, Rui Feng\textsuperscript{1,2,*}\thanks{Rui Feng is the corresponding author.}}
\address{\textsuperscript{1}School of Computer Science, Shanghai Key Laboratory of Intelligent Information Processing,\\ Fudan University, China
\\\textsuperscript{2}Academy for Engineering and Technology, Fudan University, China
\\\{jsyu19, chengy18, fengrui\}@fudan.edu.cn}
\maketitle

\begin{abstract}
Audio-visual event localization aims to localize an event that is both audible and visible in the wild, which is a widespread audio-visual scene analysis task for unconstrained videos. To address this task, we propose a Multimodal Parallel Network (MPN), which can perceive global semantics and unmixed local information parallelly. Specifically, our MPN framework consists of a classification subnetwork to predict event categories and a localization subnetwork to predict event boundaries. The classification subnetwork is constructed by the Multimodal Co-attention Module (MCM) and obtains global contexts. The localization subnetwork consists of Multimodal Bottleneck Attention Module (MBAM), which is designed to extract fine-grained segment-level contents. Extensive experiments demonstrate that our framework achieves the state-of-the-art performance both in fully supervised and weakly supervised settings on the Audio-Visual Event (AVE) dataset.
\end{abstract}
\section{Introduction}
\label{sec:intro}

\begin{figure}[t]
    \centering
    \includegraphics[width=0.47\textwidth]{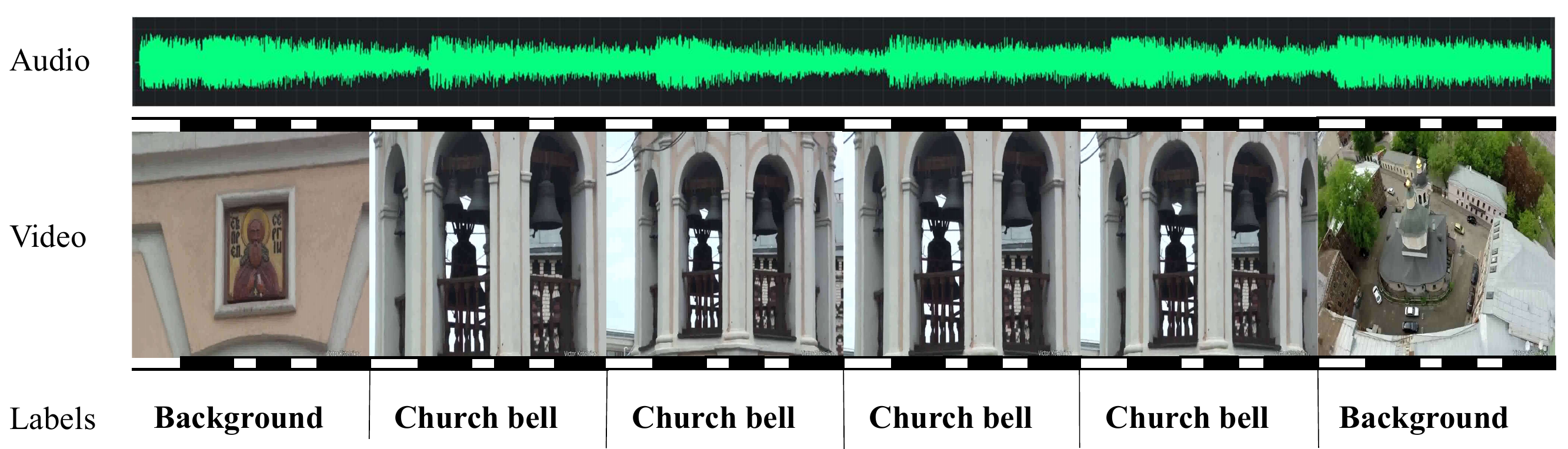}
    \caption{Illustration of audio-visual event localization task. The audio-visual event of the given video is \textit{church bell}. Segment-wise label is labeled as \textit{church bell} only when the event is both audible and visible in the current segment. Otherwise, the segment is labeled as \textit{background} so as to perform the temporal boundary localization.}
    \label{figure1}
    \vspace{-5mm}
\end{figure}

Human brains process visual and acoustics information synergistically to learn the multimodal perception and infer how to act in response. Recently, many researchers have attempted to imitate the multimodal message processing of human brains by solving various multimodal tasks~\cite{chung2017lip, gao2018learning}. However, how to localize an audio-visual event in unconstrained videos remains a challenging issue. To this end, Tian et al.\cite{tian2018audio} introduce the audio-visual event localization task, which aims to predict an audio-visual event in a video and determine the temporal boundary by distinguishing whether each video segment belongs to an event category or background. As illustrated in Fig.~\ref{figure1}, given a video of church bell, the task is to predict which temporal segment in the video contains the visual and auditory information of the \textit{church bell}, and leaves the irrelevant segments with the label \textit{background}.

When trying to understand the content of a video, it is essential to consider the video from a wider perspective to learn contextual information. To achieve this, some prior works \cite{tian2018audio, lin2019dual, ramaswamy2020makes, wu2019dual} try to obtain global contexts via self-attention mechanism \cite{vaswani2017attention} and LSTM \cite{hochreiter1997long} to enhance video comprehension. However, when generating global features, the fusion of contextual information will bring information from other segments to local content, such as introducing noise from background segments to event-related segments. Since the temporal boundary of the event is required, it is necessary to predict whether a segment belongs to the background or an event based on local contents. Therefore, though global information benefits video classification by obtaining contextual information, it also brings noisy information to local content, which has a negative effect on temporal localization.

To address this problem, we propose a novel Multimodal Parallel Network (MPN), which comprises two subnetworks to obtain global perception and local content parallelly. Similar to \cite{wu2019dual}, we decouple the task of audio-visual event localization into two subtasks: video event classification and temporal boundary localization, which are performed by two subnetworks. The classification subnetwork is designed to obtain a global representation, which is constructed by the Multimodal Co-attention Module (MCM). MCMs are cascaded in-depth to fully depict prominent characteristics. Each MCM is composed of a self-attention block and a cross-modal attention block, which can model intra- and inter-modality interactions, respectively. The localization subnetwork learns local information to predict whether a segment is event-relevant. To achieve this, we propose the Multimodal Bottleneck Attention module (MBAM), which consists of channel-wise squeeze and excitation operations~\cite{hu2018squeeze}. It can obtain segment-wise multimodal features through the squeeze operation and highlights advantageous features via the excitation operation. Besides, we also leverage the local-to-global interactions from the localization subnetwork to the classification subnetwork, so as to enrich the global representation by detailed information. Two subnetworks are trained jointly to accomplish the audio-visual event localization task. In summary, the contributions of this paper are as follows:

1. We propose to separately obtain global semantics and local contents via two parallel subnetworks, enhancing the category classification and temporal localization results.

2. We propose a Multimodal Parallel Network (MPN), which consists of a novel Multimodal Co-attention Module (MCM) and a new Multimodal Bottleneck Attention Module (MBAM) to acquire global and local features, respectively.

3. Extensive experiments show that our model can achieve the state-of-the-art results on the Audio-Visual Event (AVE) dataset both in fully and weakly supervised settings.

\section{Related Work}
\textbf{Audio-Visual Representation Learning} aims to learn discriminative audio-visual multimodal representations. Some works~\cite{aytar2016soundnet, owens2016ambient} attempt to utilize the knowledge from one modality to supervise the other modality in a “teacher-student” manner. However, the well-trained model of one modality still requires large amounts of labeled data, which restricts the generalization ability of these methods. Recently, some approaches~\cite{arandjelovic2018objects, owens2018audio, cheng2020look} leverage the correlations between audio and visual features as free supervised signals, and learn multimodal representations in a self-supervised manner. 

\noindent\textbf{Audio-Visual Event Localization} proposed by \cite{tian2018audio} requires one to identify the presence of an event that is both audible and visible in a video. Prior works focus on the inter-modality correlations and try to obtain global feature for a holistic understanding. Tian et al.~\cite{tian2018audio} use audio-guided visual attention to capture inter-modality interactions. AVDSN~\cite{lin2019dual} attempt to learn a multimodal global representation in a seq2seq manner. AVIN~\cite{ramaswamy2020makes} models the inter- and intra-modality interactions through self and collaborative attention, and learn cross-modal features via bilinear pooling. Wu et al.\cite{wu2019dual} propose a dual attention matching module to leverage global features as the reference to localize events. Though they also leverage a two-subnetwork framework, both subnetworks capture global features, thus the negative effect still remains. Different from prior works, our model perceives global semantics and unmixed local contents separately via two parallel subnetworks, thus results in better event localization performance.
\section{Method}

\begin{figure*}[t]
\centering
\includegraphics[width=0.95\textwidth]{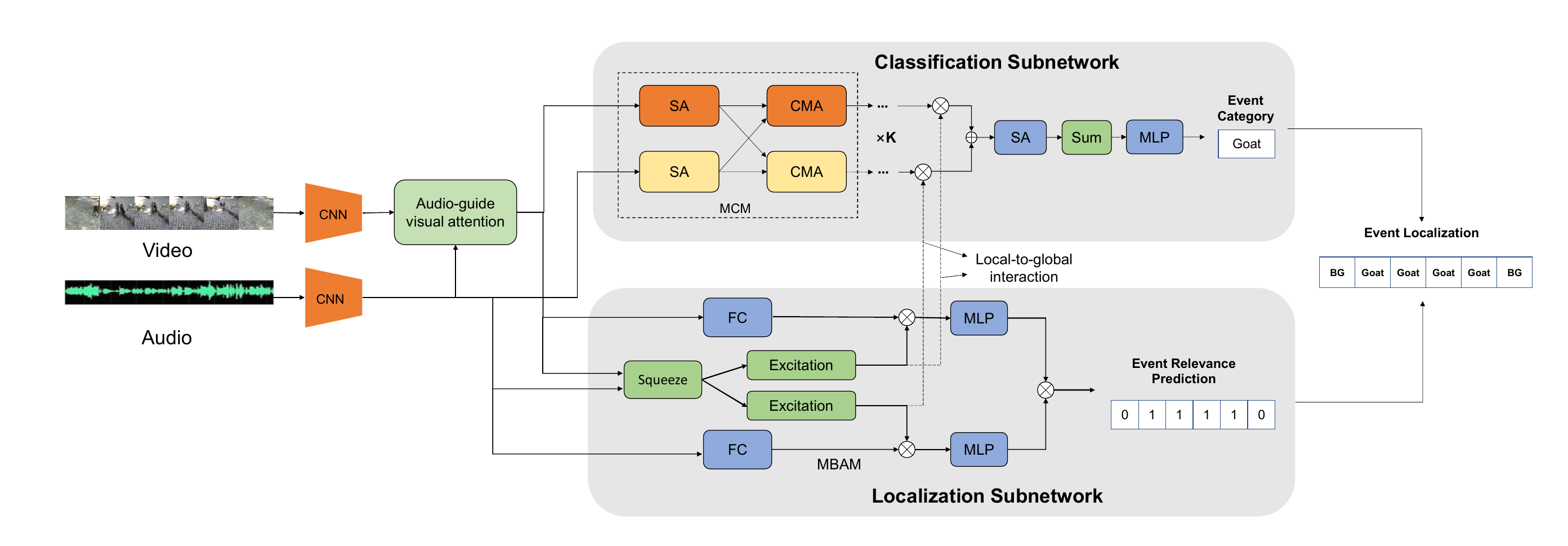}
\vspace{-3mm}
\caption{An overview of our Multimodal Parallel Network (MPN). Visual and audio features are fed into the localization and classification subnetworks simultaneously. The classification subnetwork obtains an integrated representation via the Multimodal Co-attention Module (MCM) and predicts the video-level event category. The localization subnetwork learns segment-level multimodal features by the Multimodal Bottleneck Attention Module (MBAM) and predicts event relevance labels. The predictions of two subnetworks are combined to accomplish the audio-visual event localization task.}
\label{figure2}
\vspace{-3mm}
\end{figure*}

\subsection{Preliminaries}
Audio-visual event localization aims to predict the event category and temporal boundary of an audio-visual event in an unconstrained video. Each video sequence is split into T non-overlapping segments, i.e., $V = \{v_t,a_t\}_{t=1}^T$, where $v_t$  denotes visual content in a video segment, and $a_t$ represents its corresponding audio counterpart. Following \cite{wu2019dual}, the task is decoupled into predicting two kind of scores: event-relevance scores $p_r \in \mathbb{R}^{T}$ and event category scores $p_c \in \mathbb{R}^C$, where C denotes the number of event categories. This task includes \textit{fully} and \textit{weakly} supervised event localization. In the fully supervised setting, the segment-wise labels $y_t$ are provided in the training stage, while in the weakly supervised setting, only video-level event labels are provided during training and we still need to predict segment-wise labels in the test phase.

\subsection{Multimodal Parallel Network}
We propose the Multimodal Parallel Network (MPN) for the task of audio-visual event localization, which consists of three parts: audio-guided visual attention module, classification subnetwork, and localization subnetwork, as shown in Fig.~\ref{figure2}.

\noindent\textbf{Audio-guided Visual Attention.} To capture the spatial correspondence between auditory and visual information, we adopt the audio-guided visual attention proposed by \cite{tian2018audio}. Concretely, we use a Multilayer Perceptron (MLP) with a softmax function to compute the attention weight $w_t$, then the audio-guided features $v_t^a$ are computed by: $v_t^{att} = \sum_{i=1}^k w_t^iv_t^i$.

\begin{figure}[t]
    \centering
    \includegraphics[width=0.45\textwidth]{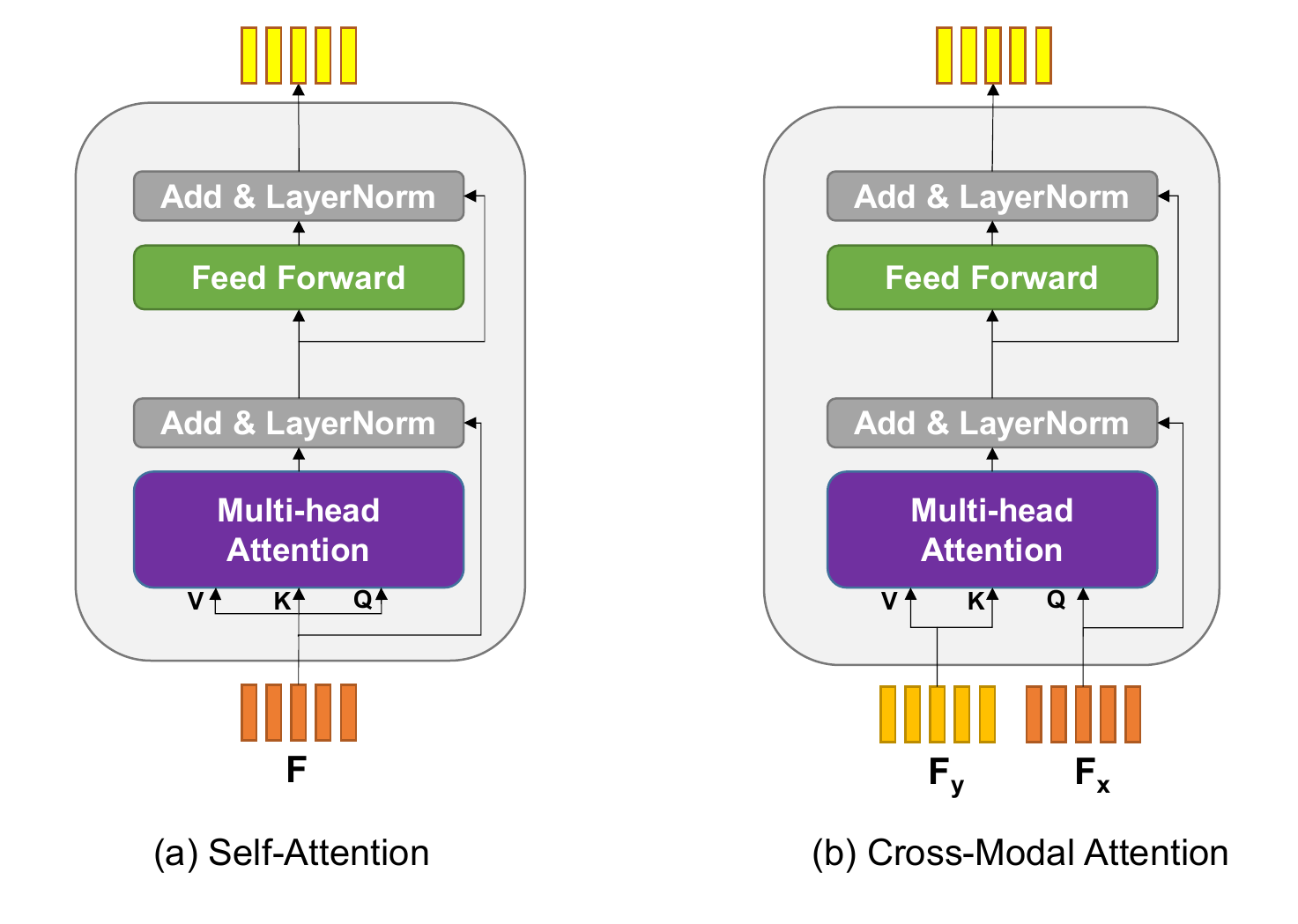}
    \vspace{-3mm}
    \caption{The structures of the Self-Attention (SA) block and the Cross-Modal Attention (CMA) block.}
    \label{figure3}
    \vspace{-3mm}
\end{figure}

\noindent\textbf{Classification Subnetwork.} The classification subnetwork obtains the global semantics and predicts event category scores. To model the inter- and intra-modality correlations and obtain global semantics, we employ the Multimodal Co-attention Module (MCM), which consists of a Self-Attention (SA) block and a Cross-Modal Attention (CMA) block. The SA block is used to model the unimodal correlation of each video segment, while the CMA block is utilized to model the correlations between different modalities. The structures of SA and CMA blocks are shown in Fig.~\ref{figure3}. We employ the similar structure of~\cite{vaswani2017attention}. Concretely, the inputs consist of queries and keys of dimension $d_k$, and values of dimension $d_v$. The dot products of each query with all keys are computed and divided by $\sqrt{d_k}$, and then applied by a softmax function. In the SA block, query, key, and value vectors are all segment-level features of the current modality while in the CMA block, the query vectors are the feature of the current modality and the key and value vectors are features from the other modality. The formulation of SA and CMA is shown as below:
\begin{equation}
    selfatt(f_t, f) = softmax(\frac{f_tf^T}{\sqrt{d_k}}f),
\end{equation}
\begin{equation}
    crossatt(v_t, a) = softmax(\frac{v_ta^T}{\sqrt{d_k}}a),
\end{equation}
\begin{equation}
    crossatt(a_t, v) = softmax(\frac{a_tv^T}{\sqrt{d_k}}v).
\end{equation}

In order to collaboratively gather information from representation subspaces at diverse positions, the multi-head attention setting is adopted, where queries, keys, and values are projected $h$ times with differing linear projections. Outputs of different heads are concatenated and then once again projected to get the final output. The multi-head attention layer is followed by a position-wise feed-forward layer, which comprises of a two-layer MLP. We also employ residual connection \cite{he2016deep} around the multi-head attention and feed-forward layer, followed by layer normalization \cite{ba2016layer}.

To further promote the learning of attention module, we refer to the insight of~\cite{chen2020dynamic} and add a temperature parameter in the softmax of the multi-head attention module, which makes attention near-uniform in the early training stage:
\begin{equation}
    S_i = \frac{exp(V_i/\tau)}{\sum_j exp(V_j/\tau)}, 
\end{equation}
where $S_i$ means the softmax output of the $i_{th}$ element in vector $V$, $\tau$ denotes the temperature parameter. 

We use a temperature annealing strategy to reduce $\tau$ from 30 to 1 in the first 10 epochs. This strategy prevents the softmax from producing near one-hot output during the early stage, which improves the performance of attention modules.

To exhaustively gather inter- and intra-modality interactions, MCMs can be cascaded in-depth. Then visual and audio features are concatenated, and an SA block is used for temporal modeling of the whole multimodal sequence. Finally, features are temporally aggregated by the max-pooling, and a two-layer MLP is used to predict the event category $p_c$ of the whole video sequence.

\begin{figure}[t]
    \centering
    \includegraphics[width=0.45\textwidth]{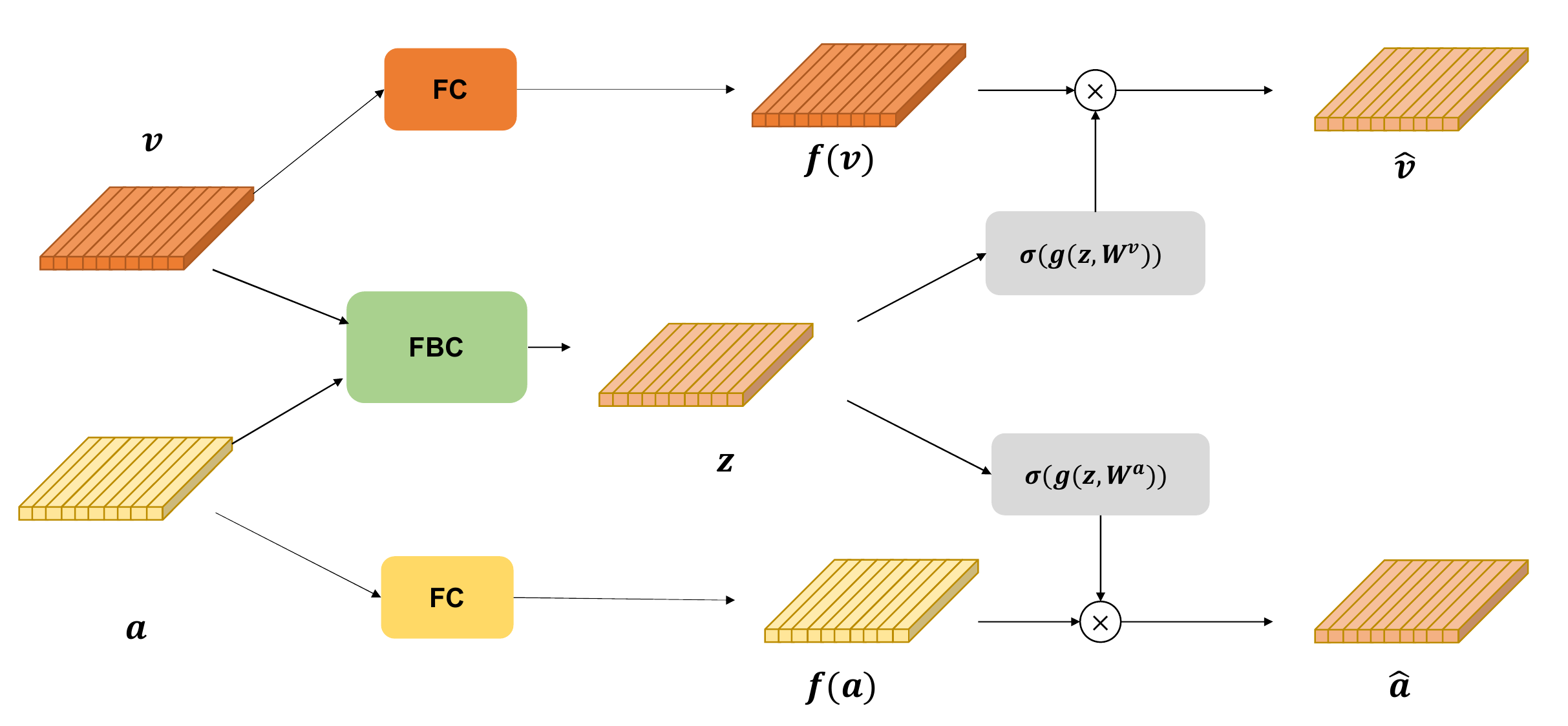}
    \caption{Overview of our Multimodal Bottleneck Attention Module (MBAM).}
    \label{figure4}
    \vspace{-3mm}
\end{figure}

\noindent\textbf{Localization Subnetwork.} The localization subnetwork learns local contents and predict event-relevant scores. This subnetwork is constructed by the Multimodal Bottleneck Attention Module (MBAM), which contains a two-step process, as shown in Fig.~\ref{figure4}. Initially, squeeze operation is conducted to integrate audio and visual features into a multimodal feature. By doing so, we can further exploit the high-level information implicit in each corresponding audio-visual segment pairs. We employ the Factorized Bilinear Coding (FBC) method \cite{gao2020revisiting} as the squeeze operation, which deals with the rank-one property of bilinear results and learns high-level multimodal information. Given visual features $\{F_t^v \in \mathbb{R}^p\}_{t=1}^T$ and audio features $\{F_t^a \in \mathbb{R}^q\}_{t=1}^T$ as inputs, FBC fuses them into a group of FBC codes $\{c_i\}_{i=1}^T$, which can be arranged in temporal order as the fused audio-visual feature $z \in \mathbb{R}^{T\times d}$, where d denotes the channel dimension of the fused feature:
\begin{equation}
\begin{cases}
    c_i' = P(\tilde{U}^\top F_t^v\circ\tilde{V}^\top F_t^a)\\
    c_i = sign(c_i')\circ max((abs(c_i')-\frac{\lambda}{2}),0),
\end{cases}
\end{equation}
where $\circ$ denotes the Hadamard product, $\tilde{U}\in \mathbb{R}^{p\times rk}$ and $\tilde{V}\in\mathbb{R}^{q\times rk}$ are learnable parameters, $\lambda$ is the regularization hyperparameter of LASSO method \cite{tibshirani1997lasso}, $r$ and $k$ are hyperparameters that denotes the rank of decomposition and number of FBC coding dictionary atoms, respectively and $P\in \mathbb{R}^{k\times rk}$ is a binary matrix, where elements in the row $l$, columns $((l-1)\times r + 1)$ to $(lr)$ are all 1, $l \in [1,k]$.

After performing the squeeze operation, multimodal feature $z$ is fused and cross-modal correlations of audio and visual segments are captured. Then an excitation operation follows to refine audio and visual features, which is implemented by a gating mechanism with a sigmoid activation:
\begin{equation}
    \phi^v = \sigma(g(z, W^v)) = \sigma(W_2^v\delta(W_1^vz)),
\end{equation}
\begin{equation}
    \phi^a = \sigma(g(z, W^a)) = \sigma(W_2^a\delta(W_1^az)),
\end{equation}
where $W_1 \in \mathbb{R}^{k_{hid}\times k}$ and $W_2 \in \mathbb{R}^{k\times k_{hid}}$ are learnable parameters, $k_{hid}$ is the dimension of hidden layer, $\delta(\cdot)$ denotes the ReLU function and $\sigma(\cdot)$ represents the sigmoid function.

We set $k_{hid}=k/2$ to form a bottleneck structure, which can constrain the complexity and increase the generation ability of the model. The excitation signal $\phi$ can be considered as channel-wise attention, which employs the high-level multimodal information to enrich unimodal features. We achieve this by the channel-wise multiplication between the excitation signal and unimodal features:
\begin{equation}
    \hat{v}_t = f(v_t) \odot \phi^v_t,
\end{equation}
\begin{equation}
    \hat{a}_t = f(a_t) \odot \phi^a_t,
\end{equation}
where $\odot$ indicates the channel-wise multiplication, and $f(\cdot)$ denotes a fully-connected layer with non-linearity.

Since audio-visual events only occur when it is both audible and visible, a two-layer MLP is applied in each unimodal stream and the final prediction is the multiplication of unimodal predictions: $p_r^{av} = p_r^a * p_r^v$.

\noindent\textbf{Local-to-Global Interactions.} In the classification subnetwork, MCM acquires global features by considering intra- and inter-modality contextual information. However, we argue that the high-level multimodal correlation inside each segment is also beneficial for global semantics. Therefore, we carried out the local-to-global interactions and leverage the gating signal of MBAM to refine the output of MCM, which is illustrated via the dashed line in Fig.~\ref{figure2}. By doing so, our classification subnetwork can leverage local content as an enhancement for more informative global semantics.

\section{Experiments}

\subsection{Experimental setup}

\noindent\textbf{Datasets.}
The Audio-Visual Event \cite{tian2018audio} dataset is a subset of the Audioset \cite{gemmeke2017audio}. This dataset consists of 4,143 videos with 28 event categories, which covers a wide range domain events such as mandolin, chainsaw, bus. Each event category contains 60 to 188 videos. Videos are all 10s clips and each video includes only one audio-visual event, which is at least 2s long. Following previous works~\cite{tian2018audio, lin2019dual, wu2019dual, ramaswamy2020makes}, we split 80\%/10\%/10\% of the dataset as the training/validation/test sets.

\noindent\textbf{Evaluation Metrics.}
Following previous works~\cite{tian2018audio, lin2019dual, wu2019dual, ramaswamy2020makes}, overall accuracy is adopted as the evaluation metric. The percentage of all matching segments in the test set is calculated as the prediction accuracy. 

\noindent\textbf{Training Strategy.}
We combine the event-relevant scores $p_r$ and event category score $p_c$ as the event localization result. Concretely, we set a threshold of 0.5 for the event relevance prediction $p_r$, which indicates the event segment if $p_r \geqslant 0.5$ and otherwise the background category.

In the fully supervised task, we use a binary cross-entropy loss for $p_r$ and a cross-entropy loss for $p_c$, and the overall objective function is as below:
\begin{equation}
    {\cal{L}} = \lambda{\cal{L}}_{BCE} + (1-\lambda){\cal{L}}_{CE}, 
\end{equation}
where $\lambda$ is a scaling parameter for balanced optimization.

It is noting that segment-level labels are only available for the fully supervised setting, thus for the weakly supervised manner, we aggregate the relevance prediction $p_r$ and event category prediction $p_c$ into a segment-level joint prediction $p_j$, and then use a Multiple Instance Learning (MIL) pooling to obtain the video-level prediction. We train our model via a multi-label soft margin loss.

\noindent\textbf{Implementation Details.}
We follow the same feature extraction method as previous works \cite{tian2018audio, lin2019dual, wu2019dual, ramaswamy2020makes}. Specifically, we employ the VGG-19 network pre-trained on ImageNet and the VGGish network pre-trained on AudioSet as the feature extraction network. Each video is splitted into several 1-second segments, and 16 frames are sampled from each segment. We extract the pool5 feature maps and use average pooling over frames to acquire the 512$\times$7$\times$7 visual features. The audio features extracted from the VGGish network are 128-D. In the training phase, we set the scaling parameter $\lambda$ as 0.6 and the learning rate as 0.0002, and train our model on a single GTX TITAN X for 300 epochs.

\subsection{Results and Discussions}

\begin{table}[t]
    \centering
    \caption{Comparisons performance (in \%) with the state-of-the-art methods in both fully supervised (Sup.) and weakly supervised (W.Sup.) manner.}
    \begin{tabular}{l|c|c}
    \hline
        Method & Sup. & W-Sup. \\ \hline
        AVE(Visual Only)~\cite{tian2018audio}& 57.4 & 53.8 \\ 
        AVE(Audio Only)~\cite{tian2018audio} & 62.3 & 57.0 \\ 
        AVE(Audio+Visual)~\cite{tian2018audio}& 72.7 & 66.7 \\ 
        ED-TCN~\cite{lea2017temporal} & 46.9 & - \\
        AVDSN~\cite{lin2019dual} & 72.8 & 66.5 \\
        DAM~\cite{wu2019dual} & 74.5 & - \\ 
        AVIN~\cite{ramaswamy2020makes} & 75.2 & 69.4 \\ \hline
        MPN (Ours) & \textbf{77.6} & \textbf{72.0} \\ \hline
    \end{tabular}

\label{table1}
\end{table}

The experimental results are shown in Table 1. To ensure a fair evaluation, we compare our method with prior works that use the same feature extraction networks, including AVEL~\cite{tian2018audio}, AVDSN~\cite{lin2019dual}, DAM~\cite{wu2019dual}, and AVIN~\cite{ramaswamy2020makes}. 

We observe that our method achieves the best performance both in the fully and weakly supervised settings on the AVE dataset. Our method outperforms the single-modality baselines proposed in~\cite{tian2018audio} and the uni-modal temporal labeling network ED-TCN~\cite{lea2017temporal} in a large margin, which proves the effectiveness of capturing audio-visual interactions. Furthermore, our MPN achieves 77.6\% accuracy in the supervised setting, which outperforms the previous best method by 2.4\%, and 72.0\% accuracy in the weakly supervised manner, which is 2.6\% higher than the previous optimal method.
\subsection{Ablation Study}

\begin{table}[t]
\caption{Ablation studies on the parallel network structure.}
    \centering
    \begin{tabular}{l|c|c}
    \hline
        Network architecture & Sup. & W-Sup. \\ \hline
        Ours w/ localization subnetwork & 74.7 & 69.2 \\ 
        Ours w/ classification subnetwork & 75.2 & 70.8 \\
        Ours w/ two subnetworks & \textbf{77.6} & \textbf{72.0} \\ \hline
    \end{tabular}
\label{table2}
\vspace{-3mm}
\end{table}

\noindent\textbf{Parallel Network Architecture.} We first investigate the effectiveness of our two subnetworks. As shown in Table 2, Ours w/ localization subnetwork means the event relevance labels and event category labels are all predicted in the localization subnetwork by two different MLPs, while ours w/ classification subnetwork means the labels are all predicted by the classification subnetwork. Results show that our parallel network design, which models global semantics and local contents separately can improve the performance on the audio-visual event localization task.

\begin{table}[t]
\caption{Ablation studies on the structure of MCM.}
    \centering
    \begin{tabular}{l|c|c}
    \hline
        MCM structure & Sup. & W-Sup. \\ \hline
        SA + SA & 74.4 & 69.2 \\ 
        CMA + CMA & 72.2 & 66.3 \\
        CMA + SA & 77.4 & 71.6 \\
        SA + CMA (Ours) & \textbf{77.6} & \textbf{72.0} \\ \hline
    \end{tabular}
\label{table3}
\end{table}

\begin{figure*}[t]
    \centering
    \includegraphics[width=0.9\textwidth]{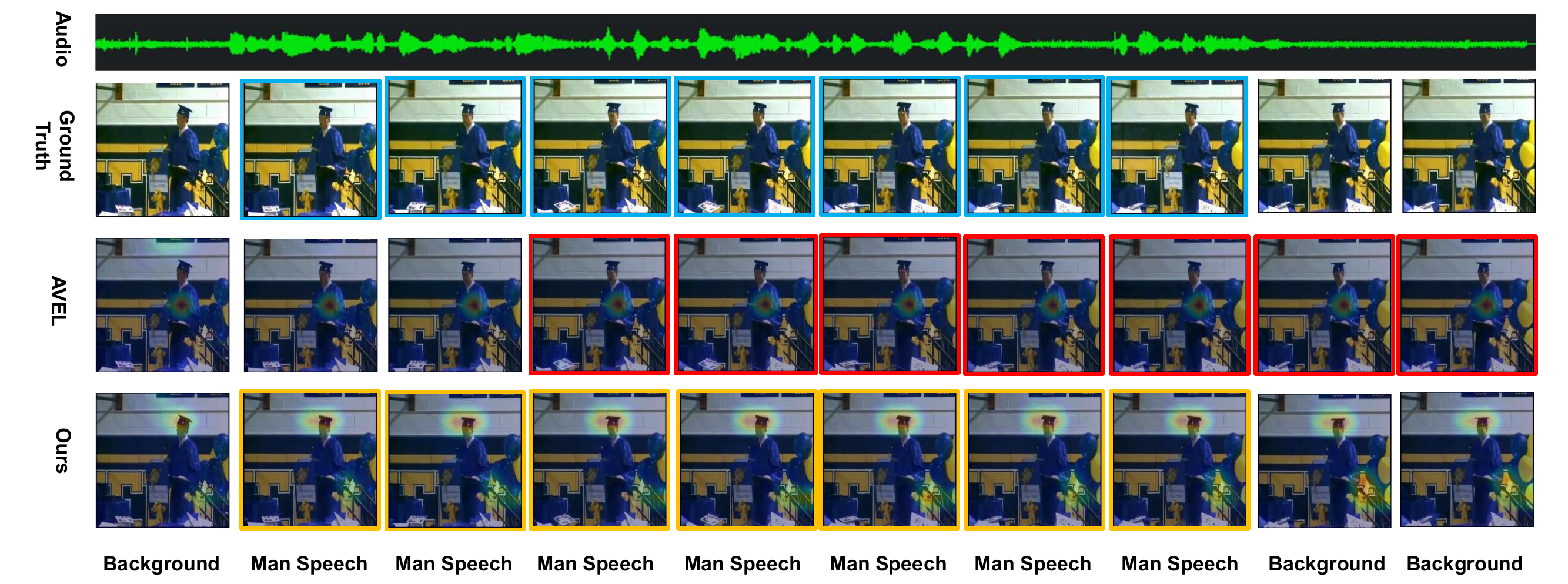}
    \vspace{-3mm}
    \caption{Qualitative results. The event label of the given video is \textit{man speech}. The blue boxes denote the ground-truth localization value, and the red boxes and the yellow boxes indicate the localization result of AVEL and our method, respectively. }
    \label{figure5}
    \vspace{-3mm}
\end{figure*}

\noindent\textbf{MCM Architecture.} We investigate the impact of the structure of MCM, as shown in Table 3. \textit{SA + SA} indicates two stacking SA blocks, while \textit{CMA + CMA} means two CMA blocks. Results show that \textit{SA + CMA (Ours)} gains better performance. This demonstrates that intra-modal information seized by SA, together with cross-modal correlations acquired by CMA are complementary to obtain the global context.

\begin{table}[t]
\caption{Ablation studies on the squeeze functions.}
    \centering
    \begin{tabular}{l|c|c}
    \hline
        Squeeze function & Sup. & W-Sup. \\ \hline
        concat & 74.8 & 68.9 \\
        element-wise product & 75.1 & 69.3 \\
        element-wise addition & 75.4 & 69.8    \\
        FBC (Ours) & \textbf{77.6} & \textbf{72.0} \\ \hline
    \end{tabular}
\label{table4}
\vspace{-3mm}
\end{table}

\noindent\textbf{MBAM Architecture.} We also study the composition of MBAM. MBAM consists of the squeeze operation and the excitation operation. we investigate the influence of different squeeze functions (shown in Table 4). Results show that the FBC is most suitable for squeeze operation, which captures the high-level multimodal features in an efficient way.

\subsection{Qualitative Results}

The visualization results are shown in Fig.~\ref{figure5}. The audio-visual event in this video is \textit{man Speech}, and the event starts from the second to the eighth video segment. Previous method AVEL predicts the second and the third segments as \textit{background} while the ground-truth is \textit{man speech}, and mispredicts the last two segments as \textit{man speech} while the ground-truth is \textit{background}. In contrast, our model obtains unmixed local contents via the localization subnetwork, and thus gives an accurate temporal localization prediction. Besides, the spatial localization result shows that our model can localize sound sources, i.e., the head of the speaker, while the AVEL mainly focuses on unrelated areas. This proves that our model can acquire more accurate video semantics.

\section{Conclusion}
In this paper, we introduce a novel Multimodal Parallel Network (MPN) for the audio-visual event localization task. Our MPN model consists of two subnetworks, one is the classiﬁcation subnetwork that predicts the category of an event, which is constructed by the Multimodal Co-attention Module (MCM) and obtains global contexts. The other one is the localization subnetwork that predicts the temporal boundary of an event, which is performed by the Multimodal Bottleneck Attention Module (MBAM) and extracts local contents. Our model can both obtain global semantics and unmixed local contents, and thus provide precise event localization results. Experiments show that our model achieves the state-of-the-art performance both in fully and weakly supervised settings on the AVE dataset.

\section{Acknowledgments}
\vspace{-2mm}
This work was supported (in part) by the Science and Technology Commission of Shanghai Municipalit (No. 20511100800 and No. 20511101502). This work was supported by the Science and Technology Major Project of Commission of Science and Technology of Shanghai (No.2021SHZDZX0103)


\begin{small}
\bibliographystyle{IEEEbib}
\bibliography{reference.bib}
\end{small}
\end{document}